\pdfoutput=1

\documentclass[11pt]{article}

\usepackage[final]{acl}

\usepackage{times}
\usepackage{latexsym}

\usepackage{graphicx}

\usepackage[T1]{fontenc}

\usepackage[utf8]{inputenc}

\usepackage{microtype}

\usepackage{inconsolata}

\usepackage{booktabs}
\usepackage{multicol}
\usepackage{multirow}
\usepackage{arydshln}
\usepackage{hyperref}
\usepackage[frozencache,cachedir=.]{minted}

\usepackage{xspace}

\newcommand{\pymarian}{\texttt{pymarian}\xspace}
\newcommand{\pymarianeval}{\texttt{pymarian-eval}\xspace}

\newcommand{\code}[1]{\texttt{#1}}
\newcommand{\hfmodel}[1]{\href{https://huggingface.co/#1}{\code{#1}}}
\newcommand*{\ditto}{\textquotedbl}

\title{PyMarian: Fast Neural Machine Translation and Evaluation in Python}

\author{Thamme Gowda$^1$ ~~~ Roman Grundkiewicz$^1$ ~~~ Elijah Rippeth$^2$  \\
{\bf Matt Post$^1$ ~~~~ Marcin Junczys-Dowmunt$^1$} 
\vspace{0.5em} \\ 
$^1$ Microsoft Translator \\ \texttt{\{thammegowda,rogrundk,mattpost,marcinjd\}@microsoft.com}
\vspace{0.25em} \\
$^2$ University of Maryland  \\ \texttt{erip@cs.umd.edu} 
  }

\begin{document}
\maketitle
\begin{abstract}
The deep learning language of choice these days is Python; measured by factors such as available libraries and technical support, it is hard to beat.
At the same time, software written in lower-level programming languages like C++ retain advantages in speed. 
We describe a Python interface to Marian NMT, a C++-based  training and inference toolkit for sequence-to-sequence models, focusing on machine translation.
This interface enables models trained with Marian to be connected to the rich, wide range of tools available in Python.
A highlight of the interface is the ability to compute state-of-the-art COMET metrics from Python but using Marian's inference engine, with a speedup factor of up to 7.8$\times$ the existing implementations.
We also briefly spotlight a number of other integrations, including Jupyter notebooks, connection with prebuilt models, and a web app interface provided with the package.
PyMarian is available in PyPI via \texttt{pip install pymarian}.
\end{abstract}

\section{Introduction}

Marian NMT\footnote{\url{https://marian-nmt.github.io}} \cite{junczys-dowmunt-etal-2018-marian} was one of the earliest training and inference toolkits for sequence-to-sequence-based machine translation.
Originally written under the name \verb|amun| and providing fast inference for Groundhog-trained models,\footnote{\url{https://github.com/pascanur/GroundHog}} it was quickly built up to also provide speedy, reliable multi-GPU and multi-node training of Transformer models, along with many other features.
It has been widely used in commercial production settings \cite{junczys-dowmunt-etal-2018-marian-cost}, for academic and industrial research, for the distribution of pre-trained models \cite{tiedemann-thottingal-2020-opus}, and as the basis for extremely fast in-browser translation \cite{bogoychev-etal-2021-translatelocally}.

Many of these features were enabled by its efficient C++-backend, but it must be admitted that this dependency is also a barrier to many researchers, who increasingly work with Python.
This paper describes a new set of Python bindings that have been added to Marian.
Written using Pybind11, these bindings are available as a \texttt{pip}-installable Python package via the Python Package Index\footnote{\url{https://pypi.org/project/pymarian}} or can be installed from Marian's source.
We then describe several features and applications facilitated by PyMarian:
\begin{itemize}
    \item \emph{Inference and training} (\S~\ref{sec:pymarianapi}).
    It is easy to load Marian-trained models and send data through them for translation.
    This also makes it easy to translate with publicly-available models, and to plug them into other Python codebases.
    \item \emph{Fast evaluation} (\S~\ref{sec:fasteval}).
    Model-based metrics such as COMET and BLEURT have demonstrated their superiority, but their provided toolsets make them slow to compute. 
    We provide \pymarianeval, which makes use of converted models, packaged in a Python CLI interface.
    \item \emph{Example applications} (\S~\ref{sec:examples}). 
    We demonstrate the versatility of \pymarian with a number of examples including a web-based demonstration framework.
\end{itemize}
A particular focus of the paper is in benchmarking popular COMET models reimplemented in Marian and available through PyMarian (\S~\ref{sec:benchmarks}), which run significantly faster than in their native implementations, providing up to 7.8x speedup in a multi-GPU setting.

\section{PyMarian API}
\label{sec:pymarianapi}

PyMarian offers \pymarian Python package containing convenient high level APIs.
We use Pybind11\footnote{\url{https://github.com/pybind/pybind11}} to bind the Python calls to Marian C++ APIs. 
\pymarian uses the same configuration system as Marian, however makes it Pythonic by offering keyword-argument (i.e., \textit{**kwargs}).

At the package's top level, we have three classes: \texttt{Translator}, \code{Trainer} and \code{Evaluator}. First two are described in this section, while the evaluator is presented in details later in Section~\ref{sec:fasteval}.

\subsection{Translator}
\label{sec:translator}
The Python API for decoding Marian models with beam search is provided by \code{Translator} class.

\begin{minted}[xleftmargin=0.2em]{python}
from pymarian import Translator
mt = Translator(
  models="model.ende.npz", 
  vocabs=["vocab.spm", "vocab.spm"]
)
hyp = mt.translate("Hello world!")
print(hyp)  # "Hallo Welt!"
\end{minted}

\noindent
It offers the same hyperparameters and functionalities as the translation service in C++, such as:
\begin{itemize}
    \item Translation speed optimization with custom beam search sizes (\texttt{beam\_size}), batch organization (\texttt{mini\_batch}, \texttt{mini\_batch\_sort}), and \texttt{fp16};
    \item $n$-best lists translation (\texttt{n\_best=True});
    \item Word alignments (e.g., \texttt{alignment="hard"}) and word-level scores (\texttt{word\_scores=True}) when more detailed subword-level information is needed (\texttt{no\_spm\_encode=True});
    \item Noised sampling from full distribution and top-K sampling with custom temperatures (e.g., \texttt{output\_sampling="topk 100 0.1"});
    \item Force-decoding of given target language prefixes (\texttt{force\_decode=True}).
\end{itemize}

\subsection{Trainer}
\label{sec:trainer}
Python API for training models supported in Marian toolkit is provided by the \code{Trainer} class.

\begin{minted}[ xleftmargin=0.2em]{python}
from pymarian import Trainer
args = {
  "type": "transformer",
  "model": "model.npz",
  "train_sets": ["train.en", "train.de"],
  "vocabs": ["vocab.spm", "vocab.spm"],
}
trainer = Trainer(**args)
trainer.train()
\end{minted}

\noindent
Complete examples are available in Marian's source code in \texttt{src/python/tests/regression}.

\section{Fast MT Evaluation in PyMarian}
\label{sec:fasteval}

The Marian NMT had been a toolkit for translation and language modeling with the emphasis on speed. 
With the recent revision of Marian toolkit, we have implemented evaluation metrics, for both training and fast inferencing, while retaining its emphasis on speed.
In addition, we have also enabled evaluator APIs in Python module, via a class named \code{Evaluator}.
   
\subsection{Evaluator}
\code{Evaluator} supports scoring MT hypothesis with either 
source, or reference, or both. 
Generally, evaluators are classified into reference-free (quality estimation) and reference-based types.
We provide implementations of both types.

\begin{minted}[xleftmargin=0.2em]{python}
from pathlib import Path
from pymarian import Evaluator

evaluator = Evaluator.new(
    model_file="marian.model.bin",
    vocab_file="vocab.spm",
    like="comet-qe", quiet=True, 
    fp16=False, cpu_threads=4)

srcs =  ['Hello', 'Howdy']
mts = ['Howdy', 'Hello']
lines = (f'{s}\t{t}' 
          for s,t in zip(srcs, mts))
scores = evaluator.evaluate(lines)
for score in scores:
    print(f'{score:.4f}')
\end{minted}

\subsection{Metrics}
Along with providing implementation for Evaluator, we also provide checkpoints for some of the popular MT metrics, such as COMETs and BLEURT.
Since the checkpoint file format of the existing metrics are incompatible with Marian toolkit, we have converted them to the required format and released on Huggingface.\footnote{\url{https://huggingface.co/models}}
Table~\ref{tab:models} shows the available models and their IDs on HuggingFace hub.

\begin{table*}[ht]
\small
\centering
\begin{tabular}{l l c  l} 
\toprule
\textbf{Metric} & \textbf{Fields}  & \textbf{Reference} & \textbf{HuggingFace ID} \\ 
\midrule \midrule
bleurt-20 & T, R & \citet{sellam-etal-2020-bleurt}  & \hfmodel{marian-nmt/bleurt-20} \\
wmt20-comet-da & S, T, R & \citet{rei-etal-2020-unbabels} & \hfmodel{unbabel/wmt20-comet-da-marian}  \\
wmt20-comet-qe-da & S, T & \ditto & \hfmodel{unbabel/wmt20-comet-qe-da-marian} \\
wmt20-comet-qe-da-v2 & S, T & \ditto & \hfmodel{unbabel/wmt20-comet-qe-da-v2-marian}\\
wmt21-comet-da & S, T, R & \citet{rei-etal-2021-references} & \hfmodel{unbabel/wmt21-comet-da-marian}   \\
wmt21-comet-qe-da & S, T & \ditto  & \hfmodel{unbabel/wmt21-comet-qe-da-marian}\\
wmt21-comet-qe-mqm & S, T & \ditto & \hfmodel{unbabel/wmt21-comet-qe-mqm-marian} \\
wmt22-comet-da & S, T, R & \citet{rei-etal-2022-comet}& \hfmodel{unbabel/wmt22-comet-da-marian} \\
wmt22-cometkiwi-da & S, T & \citet{rei-etal-2022-cometkiwi} & \hfmodel{unbabel/wmt22-cometkiwi-da-marian}  \\
wmt23-cometkiwi-da-xl & S, T & \citet{rei-etal-2023-scaling} & \hfmodel{unbabel/wmt23-cometkiwi-da-xl-marian}  \\
wmt23-cometkiwi-da-xxl & S, T & \ditto &\hfmodel{unbabel/wmt23-cometkiwi-da-xxl-marian} \\ 
cometoid22-wmt21 & S, T & \citet{gowda-etal-2023-cometoid} & \hfmodel{marian-nmt/cometoid22-wmt21} \\
cometoid22-wmt22 & S, T & \ditto & \hfmodel{marian-nmt/cometoid22-wmt22} \\
cometoid22-wmt23 & S, T & \ditto & \hfmodel{marian-nmt/cometoid22-wmt23} \\
chrfoid-wmt23 & S, T & \ditto & \hfmodel{marian-nmt/chrfoid-wmt23}\\
\bottomrule
    \end{tabular}
\caption{List of metrics supported in \pymarian, their required fields, reference, and HuggingFace model IDs.
Fields S, T, and R are \emph{source}, \emph{translation} (also variously called the \emph{candidate} or \emph{hypothesis}), and \emph{reference}, respectively.}
\label{tab:models}
\end{table*}

Using the \code{Evaluator} API, we have developed a command-line utility named \code{pymarian-eval}, which internally takes care of downloading models from HuggingFace model hub and caching them locally.

We provide \code{-a|----average} option for obtaining the system level score only (\code{-a only}), segment level scores only (\code{-a skip}), or both where average is appended  (\code{-a append}). For example, 
\begin{minted}{shell}
pymarian-eval -m wmt22-cometkiwi-da \
 -s src.txt -t mt.txt -a only
\end{minted}

The current toolkits that originally implement the popular metrics consume higher memory and time for loading the checkpoints than necessary. 
This is increasingly problematic as metric checkpoint files are getting bigger over the years. 
The format used by Marian is optimized for faster loading with minimal memory overhead.
We present the model loading time and memory utilization in Table~\ref{tab:warmup-stats}. 
For instance, consider \code{wmt23-cometkiwi-da-xl}, whose checkpoint file is 13.9GB.\footnote{\code{wmt23-cometkiwi-da-xxl} is 42.9GB and we were unable to load it on the GPUs used for benchmarks in this paper (32GB V100).} 
The \textit{original} tool (comet-score) takes 27GB~RAM and 530~seconds to warmup on 8 GPUs, where as \code{pymarian-eval} achieves the same in half the RAM and only 12~seconds.   

\begin{table*}[hbt]
\centering
\small \renewcommand{\arraystretch}{1.2}
\begin{tabular}{l rrr rrr | rr rr } 
\toprule
 & \multicolumn{6}{c |}{\textbf{Time (seconds)}} & \multicolumn{4}{c}{\textbf{Memory (MB)} } \\
  & \multicolumn{3}{c}{\textbf{1 GPU}} & \multicolumn{3}{c |}{\textbf{8 GPUs}} & \multicolumn{2}{c}{\textbf{1 GPU}} & \multicolumn{2}{c}{\textbf{8 GPUs}} \\ \cline{2-11}
\textbf{Model} & \textbf{Orig} & \textbf{Ours} & \textbf{Speedup} & \textbf{Orig} & \textbf{Ours} & \textbf{Speedup} & \textbf{Orig} & \textbf{Ours} & \textbf{Orig} & \textbf{Ours} \\ 
\midrule
bleurt-20     & 23.7 & 3.0 & 7.9x   & NA & 8.4 & NA & 6,606 & 2,640 &  NA & 3,455 \\
wmt20-comet-da & 37.0 & 4.6 & 8.0x & 193.8 & 9.7 & 19.9x & 5,387 & 2,782 & 5,388 & 3,598 \\
wmt20-comet-qe-da & 32.6 & 3.8 & 8.6x & 197.3 & 8.9 & 22.1x & 5,276 & 2,682 & 5,278 & 3,499 \\
wmt22-comet-da    & 37.9 & 4.5 & 8.5x & 193.5 & 9.7 & 20.0x & 5,365 & 2,786 & 5,364 & 3,603 \\
wmt22-cometkiwi-da & 33.9 & 3.3 & 10.2x & 199.1 & 8.8 & 22.7x & 5,244 & 2,623 & 5,246 & 3,438 \\
wmt23-cometkiwi-da-xl & 108.5 & 7.5 & 14.4x & 530.2 & 12.1 & 43.9x & 27,554 & 13,815 & 27,554 & 14,631 \\
\bottomrule
\end{tabular}
    \caption{Model load time (seconds) and memory (megabytes) taken to initialize the models and score a single example. 
Marian and \pymarian use memory-mapped files, which enable faster loading than original implementation.
Numbers are the average of three runs. }
    \label{tab:warmup-stats}
\end{table*}

\begin{table*}[htb]
\centering 
\small
\begin{tabular}{ l  r  r  r  r  r  r  r } 
\toprule
& \multicolumn{4}{c}{\textbf{ Time (seconds)}}  & \multicolumn{3}{c}{\textbf{ Speedup}} \\
\textbf{ Metric} & \textbf{Original} & \textbf{Marian} & \textbf{PyM} & \textbf{PyM FP16} & \textbf{Marian} & \textbf{PyM} & \textbf{PyM FP16} \\
\midrule \midrule
\multicolumn{8}{c}{\textit{1 GPU}} \\
\midrule
bleurt-20 & 2312$\pm$2.2 & 635$\pm$0.3 & 656$\pm$0.3 & 467$\pm$0.6 & 3.6x & 3.5x & 4.9x \\
wmt20-comet-da & 3988$\pm$0.8 & 954$\pm$1.0 & 968$\pm$4.7 & 783$\pm$5.1 & 4.2x & 4.1x & 5.1x \\
wmt20-comet-qe-da & 2529$\pm$0.4 & 608$\pm$3.7 & 623$\pm$3.6 & 501$\pm$0.3 & 4.2x & 4.1x & 5.0x \\
wmt22-comet-da & 3772$\pm$1.3 & 858$\pm$4.6 & 884$\pm$4.5 & 676$\pm$0.8 & 4.4x & 4.3x & 5.6x \\
wmt22-cometkiwi-da & 2357$\pm$2.0 & 419$\pm$0.4 & 437$\pm$1.7 & 327$\pm$1.0 & 5.6x & 5.4x & 7.2x \\

wmt23-cometkiwi-da-xl & 17252$\pm$0.7 & 3405$\pm$4.7 & 3480$\pm$3.9 & 1949$\pm$3.1 & 5.1x & 5.0x & 8.8x \\
\midrule \midrule
\multicolumn{8}{c}{\textit{8 GPUs}} \\
\midrule
bleurt-20 & NA & 85$\pm$0.1 & 99$\pm$0.1 & 76$\pm$0.4 & NA & NA & NA \\
wmt20-comet-da & 926$\pm$1.0 & 125$\pm$0.1 & 146$\pm$0.7 & 124$\pm$1.0 & 7.4x & 6.3x & 7.5x \\
wmt20-comet-qe-da & 622$\pm$0.1 & 82$\pm$0.1 & 95$\pm$0.2 & 81$\pm$0.2 & 7.6x & 6.5x & 7.7x \\
wmt22-comet-da & 896$\pm$0.8 & 114$\pm$0.1 & 135$\pm$0.3 & 111$\pm$0.7 & 7.8x & 6.6x & 8.1x \\
wmt22-cometkiwi-da & 562$\pm$0.7 & 59$\pm$0.1 & 72$\pm$0.1 & 58$\pm$0.1 & 9.5x & 7.8x & 9.6x \\
wmt23-cometkiwi-da-xl & 3288$\pm$1.8 & 662$\pm$2.6 & 862$\pm$13.3 & 258$\pm$0.7 & 5.0x & 3.8x & 12.7x \\
\bottomrule
\end{tabular}
\caption{Time taken (seconds) to score the benchmark datasets having 364,200 examples, and the speedup of our implementation with respect to the original. 
Numbers are the average of three runs on one and eight GPUs. PyM is short for PyMarian.
The column with FP16 is half-precision, and the rest are full-precision (32-bit).}
\label{tab:benchmarks}
\end{table*}

\begin{table*}[htb]
\centering \small
\begin{tabular}{ l | r | r | r | r | r }
\toprule
 & \multicolumn{3}{c|}{\textbf{Score}}& \multicolumn{2}{c}{\textbf{Error}} \\
\textbf{Metric} & \textbf{Original} & \textbf{Marian FP32} & \textbf{Marian FP16} & \textbf{Marian FP32} & \textbf{Marian FP16} \\
 \midrule
bleurt-20 & 0.7255 & 0.7252 & 0.7211 & 0.0003 & 0.0044 \\
wmt20-comet-da & 0.5721 & 0.5720 & 0.5716 & 0.0001 & 0.0005 \\
wmt20-comet-qe-da & 0.1933 & 0.1932 & 0.1924 & 0.0001 & 0.0009 \\
wmt22-comet-da & 0.8462 & 0.8461 & 0.8427 & 0.0000 & 0.0034 \\
wmt22-cometkiwi-da & 0.7984 & 0.7984 & 0.7981 & 0.0000 & 0.0003 \\
wmt23-cometkiwi-da-xl & 0.6840 & 0.6839 & 0.6862 &  0.0001 & 0.0023\\
\bottomrule
\end{tabular}
\caption{The average scores produced by the original implementation and ours.  
The columns named `Error' are the absolute difference between the average of scores from  the original and our implementations.}
\label{tab:err-margin}
\end{table*}

\subsection{Benchmarks}
\label{sec:benchmarks}

A concern with new implementations is the risk of producing incompatible results.
We therefore compare our model conversion and implementations carefully so as to ensure that \pymarianeval produces the same results. 

Our benchmark setup is as follows: 
\begin{itemize}
    \item Dataset: WMT23 General Translation submissions; we combine all systems for all languages pairs, which results in a total of 364,200 examples.
    \item COMETs original implementation: unbabel-comet v2.2.2; transititive dependencies: torch v2.4.0, pytorch-lightning v2.3.3, transformers v4.43.3
    \item BLEURT original implementation is installed from source repository\footnote{\url{https://github.com/google-research/bleurt/tree/cebe7e6f}}; transititive dependencies: tensorflow v2.17.0
    \item Marian v1.12.31, compiled with GCC v11.
    \item Python v3.10.12, Ubuntu 22.04.3, on Intel(R) Xeon(R) Platinum 8168 CPU @ 2.70GHz
    \item GPU: 8x Nvidia Tesla V100 (32GB); Driver v525.105.17, CUDA v12.3
    \item Batch size is 128, except for wmt23-cometkiwi-xl, the largest batch size that worked are: 64 for eight GPUs and 32 for one GPU.
\end{itemize}

In Table~\ref{tab:benchmarks}, we report the time taken by original toolkits (Pytorch based comet-score and Tensorflow based bluert) and our implementation. 
For ours, we report Marian (binary produced by C++), and \code{pymarian-eval}(with float32 and float16 precisions).
In addition, we also present the average of segment scores, and error, i.e., the absolute difference between the scores produced by the original and ours.
The scores and derived errors for our implementation remain consistent regardless of whether the C++ implementation is invoked via the command line binary (marian evaluate) or through the Python bindings wrapper (pymarian-eval). Additionally, the scores are identical whether the benchmarks are conducted on a single GPU or parallelized across multiple GPUs. 
We avoid repetition, and instead present only the values for full-precision (FP32) and half-precision (FP16). 
As shown in Table~\ref{tab:err-margin}, ours yield the same scores as the original, with minor discrepancies attributable to floating-point calculations.

In addition to providing significantly faster processing times, \pymarianeval provides a flexible CLI tool with a natural POSIX interface (e.g., STDIN/STDOUT, use of TSV formats).
This allows it to integrate well with other tools, such as SacreBLEU's test-set downloading capabilities\cite{post-2018-call}.

\section{Example applications}
\label{sec:examples}
A Python API makes it simple to incorporate Marian models into the many Python-native settings that researchers are accustomed to.
In this section we illustrate example use cases and applications of PyMarian, demonstrating its versatility.

\subsection{Jupyter notebook}
PyMarian makes it easy to use Marian-trained models in interactive sessions such as Jupyter Notebook-like\footnote{\url{https://jupyter.org}} environments.
We provide an example notebook for translation, training, and evaluation via Google Colab at \url{https://colab.research.google.com/drive/1Lg_W5K2nLtvaKfLuHjc-LAajenI_SGL3}

\subsection{OPUS-MT models}
Over the years, Marian NMT has been widely adopted by the community to train and release open-sourced machine translation systems. 
One of the largest projects developing such resources is OPUS-MT, which offers over 1,000 pre-trained models \cite{TiedemannThottingal:EAMT2020,tiedemann2023democratizing}.
PyMarian provides a seamless interface to decode with these existing Marian-trained models.

\subsection{Web app demo}

\begin{figure}
    \centering
    \includegraphics[width=\linewidth]{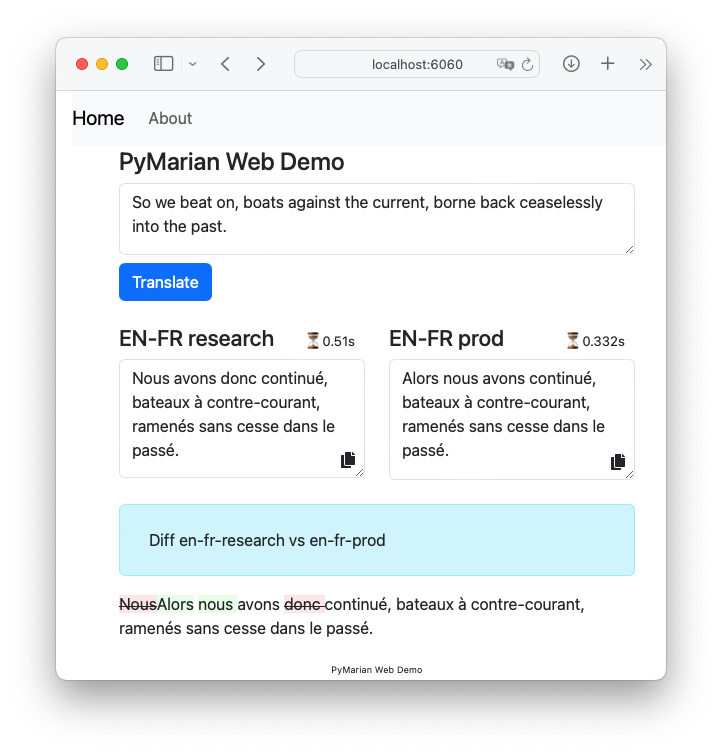}
    \caption{PyMarian web demo with two outputs and diff between them.}
    \label{figure:demo}
\end{figure}

\pymarian permits easy connection from Marian models to Python's visualization libraries.
We incorporate a Flask-based web server that can display a range of models side by side.\footnote{\url{https://github.com/marian-nmt/pymarian-webapp}}
It supports loading of models from local disk (type ``base'') or connecting to Microsoft's API (type ``mtapi'').
\begin{minted}{yaml}
translators:
  en-de-research:
    type: base
    name: research
    model: /path/to/marian.npz
    vocab: /path/to/vocab.spm
  en-de-prod:
    type: mtapi
    name: prod
    subscription-key: {redacted}
    source-language: en
    target-language: de
\end{minted}
\noindent Figure~\ref{figure:demo} provides an example of this interface.
Due to the flexibility of Python, extending the model to support other types is simple. 


\section{Related Work}

A wide range of Python toolkits exist for training and inference for the ``classical'' (i.e., not LLM-based) sequence-to-sequence approach to machine translation.
One of the most popular is Meta's Fairseq \cite{ott-etal-2019-fairseq}, which supports a wide range of training and inference features, including multi-GPU and multi-node training.
Amazon's Sockeye \cite{hieber-etal-2022-sockeye3} is another option; while it has fewer features than fairseq, it is known for its strong software engineering practices and flexibility.
Both of these toolkits are based on Pytorch \cite{paszke-etal-2019-pytorch}, and support research and production use cases.
Sockeye has recently (as of June 7, 2024) been end-of-lifed.\footnote{\url{https://github.com/awslabs/sockeye/commit/e42fbb30be9bca1f5073f092b687966636370092}}

A significant amount of research and development activity takes place using HuggingFace's popular \verb|transformers| package.
Work in this area tends to be much more research-focused, however, which means that software-engineering practices and speed are sacrificed in favor of rapid development.
HuggingFace also provides a data store for a huge range of datasets and models.
VLLM is a recent project that provides fast, production-oriented inference for HuggingFace models \cite{kwon-etal-2023-efficient}.

There is support for loading Marian models in HuggingFace, largely provided by \cite{tiedemann-thottingal-2020-opus}.
However, not all Marian model features are supported.
\pymarian provides Python-based access to any Marian model, with C++ inference speeds.

\section{Summary}

We have introduced \pymarian, a set of Python bindings that export Marian's fast training and inference capabilities to Python settings, without requiring any model conversion into much slower frameworks.

These bindings enable a range of integrations with Python---the preferred toolkit for research in NLP and MT---making available Marian's high training and inference speeds.
In particular, it enables \pymarianeval, an implementation of COMET and BLEURT models yielding speedups as high as 
7.8x (for wmt22-cometkiwi-da) on eight GPUs, and never less than $3.5$x. 
\pymarianeval is also significantly faster at loading models, and up to 44x (for wmt23-cometkiwi-da-xl) on eight GPUs. 
These models are made available on Hugginface and are seamlessly downloaded at runtime.

\section*{Limitations}

PyMarian aims to enhance the accessibility and usability of Marian NMT and publicly available machine translation models trained with the toolkit.
The primary limitation of PyMarian is that it is designed specifically for Marian-trained models, which may restrict its flexibility for users who wish to integrate models trained using other frameworks or custom architectures.
Additionally, we have implemented only the most popular evaluation metrics, such as COMET and BLEURT, which may not encompass all the evaluation metrics required for specific research or application needs.

COMET-Kiwi models require users to accept a custom license and terms of use.
To ensure that the license is preserved in the Marian-trained versions, we collaborated with the original authors.
They now host our models exclusively on HuggingFace, where users must accept the same license before downloading.
The availability of these models is subject to their decisions.

The reported benchmarks are based on specific hardware and software settings and may not fully capture the variability in real-world scenarios. 
Despite the optimizations, running MT evaluation metrics can be resource-intensive, requiring significant computational power. 
This limitation may pose challenges for users with limited access to high-performance computing resources.

Finally, as with any open-source project, the long-term maintenance and support of PyMarian depend on the community's contributions and engagement.
Ensuring the project's sustainability requires continuous collaboration and support from the community.



\bibliography{anthology,custom}


\end{document}